\documentclass[11pt,a4paper]{article}
\usepackage[hyperref]{acl2019}
\usepackage{times}
\usepackage{latexsym}

\usepackage{url}
\usepackage{graphicx}
\usepackage{float}
\aclfinalcopy 

\title{Immersive Text Game and Personality Classification}

\author{Wanshui Li, Yifan Bai, Jiaxuan Lu, Kexin Yi 
\\University College London}

\date{}

\begin{document}
\maketitle
\begin{abstract}
We designed and built a game called \textit{Immersive Text Game}, which allows the player to choose a story and a character, and interact with other characters in the story in an immersive manner of dialogues. The game is based on several latest models, including text generation language model, information extraction model, commonsense reasoning model, and psychology evaluation model. In the past, similar text games usually let players choose from limited actions instead of answering on their own, and not every time what characters said are determined by the player. Through the combination of these models and elaborate game mechanics and modes, the player will find some novel experiences as driven through the storyline.
\end{abstract}

\section{Introduction}

Text generation has become increasingly popular with the development of natural language processing. Several applications are designed with this technology, such as generating stories from keywords, and text adventure game. People have witnessed great success in this area as the models have show promising ability to write a story after training on numerous texts.

A text adventure game called AI Dungeon 2 \cite{walton} was released in 2019, built on the OpenAI \textbf{GPT-2} model \cite{radford2019language}. As the game starts, the player can choose a specific style of the story, such as fantasy or zombie, and the identity of the player is also chosen. Then, the game begins with a randomly generated text which specifies the background of this adventure. The player can type their actions without restrictions at each round, and the game will automatically process a response describing the following plots. This adventure generated by artificial intelligence (AI) constructs infinitely possible stories with coherence to some extent. 
\\\\
Mainly motivated by AI Dungeon 2, our Immersive Text Game is a dialogue-type text game based on famous TV series and film scripts. The player will choose a specific script, a character within and a topic (e.g. politics, art) from a selected range at the start. Once the game begins with a starting text excerpted from the original script, the player can follow the starting text with a line of user input. The model based on the GPT-2 \textit{Plug and Play Language Model} (\textbf{PPLM}) will then generate the following story until the chosen character appears in the dialogue. In this way, the player can play the role of the chosen character, and interact by chatting with other characters in the dialogue. To keep the completeness and relevance of the whole dialogue, the model will generate the following story regarding the chosen topic, keywords extracted from the original scripts. Moreover, we also combine the causes, effects, and attributes of the player's responses to ensure the logic of the generated texts. This is introduced by the nine if-then relationships provided by \textit{COMmonsEnse Transformers} (\textbf{COMET}) commonsense reasoning model \cite{bosselut2019comet}. 

Furthermore, we suppose that the texts input and actions taken by the player will have an implication in the personality and psychological conditions to some extent, if the player plays as if he/she were the character. Hence, while enjoying the immersive experience in our text game, the player can also investigate their personality through the game. We consider the \textbf{MBTI} (i.e. Myers-Briggs Type Indicator) dataset to train our classification model. The \textbf{BERT} classification model \cite{devlin2018bert} uses all the input of the player in the game. And it will conduct text classification which evaluates the personality and psychological condition of the player. The accuracy of the BERT model is above 0.7 for most personality categories.

\section{Related Work}
The existing works related to the design and implementation of the Immersive Text Game can be divided into two categories: text adventure games and their related technologies.\\
\\
\textbf{Text Adventure Games.} Tabletop Role-Playing Game (TRPG) is a kind of traditional text or speech adventure game. \cite{TRPG} For instance, \textit{Dungeon\&Dragons}, \textit{Call of Cthulu} are typical games of this category. TRPG usually involves two human-controlled roles: a game master and players. The players describe the actions of their characters by text. The game master controls the process of the game and gives the response to the actions that players take. People have considered using AI to play the part of the game master, while no AI at that time is plausible due to the complexity of the story.

As the methods on text generation and machine commonsense develop, AI Dungeon was released, which involved AI to be the role similar to the game master. Compared to TRPG, it used a simpler storytelling scheme instead of complex inference. AI based on the model GPT-2 can provide reasonable responses corresponding to the action that players type \cite{radford2019language}. 

On the other hand, people also considered to construct an AI agent to play the text adventure games, such as interactive fiction games \cite{Kostka_2017}. An AI agent called Golovin was trained using Long short-term memory (\textbf{LSTM}) \cite{LSTM} and playing algorithm based on command generators. Golovin had been tested for several text adventure games with satisfactory results.\\
\\
\textbf{Text Generation.} The main method used in AI text adventure games is text generation. Traditional natural language generation models use LSTM or Recurrent neural network (\textbf{RNN}) as the encoder and greedy search, beam search or sampling method as the decoder. In 2014, Sutskever designed a conditional sequence to sequence language model based on LSTM and argmax function \cite{sutskever2014sequence}. However, for long sentences, this model showed a problem of long term dependency and loss of information. To handle this problem, attention was introduced in the mechanism in the same year \cite{bahdanau2014neural}. To optimize the speed of the model, google research team proposed an attention-based transformer architecture instead of using RNN. This model improved the BLUE scores for plenty of existing models \cite{vaswani2017attention}.

Then, a new model GPT-2 became popular after 2019 with its application, text adventure game \textit{AI dungeon 2}. GPT-2 was a language model based on transformer with a few modifications on layer normalization and batch size compared to the original GPT \cite{radford2019language}. However, the output of GPT-2 exposed a certain degree of randomness thematically. Thus, in order to control the topic of text generation without the significant cost of modifying the original model, fine-tuning and re-training, plug and play language models(i.e. PPLM) for the conditional sequence to sequence generation was introduced by Uber AI \cite{dathathri2019plug}. In this architecture, the bag of words pre-specified by users was used to switch the topics of the generated text.\\
\\
\textbf{Commonsense Reasoning.} Commonsense reasoning models generate the potential effects, intentions, and attributes of an event described by text. One existing commonsense knowledge graph called \textit{ConceptNet} \cite{speer2016conceptnet} provided inference relations in the given sentences, such as causes and motivations. In terms of word relations, the accuracy of ConceptNet 56.1\% outperformed other systems.


Another popular commonsense if-then knowledge graph was the \textit{ATlas Of MachIne Commonsense} (\textbf{ATOMIC}) \cite{sap2018atomic}. The inference of ATOMIC was event-based, which provided nine types of hierarchical relations, mainly classified into ``causes'', ``effects'' and ``stative''. The model used GloVe pre-trained embeddings combined with ElMO as the encoder and a unidirectional gated recurrent unit as the decoder. The evaluation of this model was conducted by the average BLUE score and human evaluation, which showed an 86.2\% accuracy of the relationship description.

Based on the two knowledge graphs above, COMET \cite{bosselut2019comet} made a combination of ConceptNet and ATOMIC dataset. COMET learned existing tuples to construct new tuples in the commonsense knowledge base by large transformer language models. The performance of COMET by human evaluation was promising with 77.5\% valid descriptions of ATOMIC and 91.7\% valid descriptions of ConceptNet.\\
\\
\textbf{Keywords Extraction} The general process to extract information from the text were tokenization, pos-tagging, entity detection such as noun phrase chunking. The common way to implement this process was to use the Natural Language Toolkit (NLTK) in Python as it provided models for tokenization, stemming, and parsing \cite{NLTK}. However, popular algorithms such as Cocke–Younger–Kasami algorithm (CYK) and Conditional random fields \cite{sutton2010introduction} for parsing exposed computational problems.
\\ Another way to extract keywords from sentence was to use bigram taggers and semi-context-free grammar matching rules \cite{Babluki_2013}. This model ran faster than the default NLTK method.\\
\\
\textbf{Text Classification.} Text classification could be performed by algorithms such as Naive Bayes, SVM, and neural networks. Naive Bayes \cite{raschka2014naive} was a traditional algorithm for classification problems in machine learning. The core idea of Naive Bayes in text classification was the Bayes theorem, bag of word representations, and conditional independence assumptions. The classifier will classify the documents into the class with maximum likelihood.

In 2018, bidirectional encoder representations from transformers (BERT) \cite{devlin2018bert} was designed along with its desired applications. The pre-trained BERT model was a masked language model based on a deep bidirectional transformer. After fine-tuning, BERT with softmax function could also be used for text binary classification \cite{sun2019finetune}.

\section{Methods}
\begin{figure}[ht]
\centering
\includegraphics[scale=0.4]{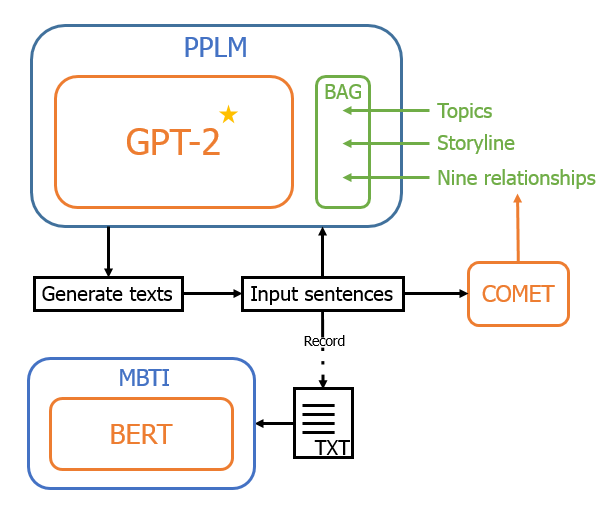}
\caption{The pipeline of the Immersive Text Game.}
\label{Frame}
\end{figure}

In order to build the Immersive Text Game, our project, therefore, consists of three parts: PPLM, COMET, and personality classification model based on Myers-Briggs Type Indicator (MBTI). Figure \ref{Frame} shows the pipeline behind the Immersive Text Game.

\subsection{Plug and Play Language Model}
\subsubsection{Data}

There are two parts of the data used in the project. One is the scripts of the shows, which in our case are \textit{Friends} and \textit{Sherlock Series 1}. The other set of data are the summaries of the shows retrieved from Wikipedia, on these two series respectively. We use them to conclude keywords from the stories, which are then put into the bag of words used in PPLM for the directed generation of text. 

\subsubsection{Training Process}
The main steps consist of four parts, two highlights, and four merits.\\
\\
\textbf{Fine-tuning the GPT-2 Model.} In order to tailor to the format of the script and the characters of a specific show, the pre-trained GPT-2 model by OpenAI was fine-tuned based on the script of the show. As the process was done on Google Colaboratory (Colab), which can only provide one GPU with the model up to Nvidia P100 with 16GB memory, we used the medium size GPT-2 model with 355 million parameters.

It can be any scripts, but there is only one requirement that the sentences should be: “Name: sentences”, for example: “Monica: There's nothing to tell! He's just some guy I work with!” This requirement can better help us to figure out the people who are talking so that we achieve the 1st highlight to let the player select the character they want to be.\\
\\
\textbf{Keywords Extraction.} To push the development of the plot, we extract the keywords or phrases from the script. Because our script file is normally quite large, we find a quick and efficient way \cite{Babluki_2013} to extract rather than simply achieve it with NLTK. The words or phrases we extract basically are noun or verb, and it does not use parsing like CYK algorithm due to the high complexity of $O(N^3)$. Though parsing is a better way to conduct extraction, an approach to process long texts in a short time would be practical.

There are three schemes to extract information: directly extracting from the whole script, extracting from each season, or extracting from every episode. To perform efficiently, we choose the second one. Moreover, for each season, we crawl summaries for each episode from Wikipedia and put them into one file for extracting seasonally, and it can improve the accuracy in some ways. In order to push plot development, we set a point to change the keywords, for example, when the player inputs five times, we change the key words file to next season’s file. But how can we make use of them?\\
\\
\textbf{Controlling State-of-art Models.} There was a hot discussion about “how high-level a language model is, and whether it’s harmful?” Introduced by OpenAI, the GPT-2 model has slowly cooled down, and the practicality problem of language models has become increasingly apparent: The output of a language model may be challenging to find its practical application, and can only be regarded as a scoring tool. For instance, when we input the beginning texts, these models can generate long texts with proper spelling, grammar, and other satisfying linguistics; however, we cannot control them to show particular styles, topics or logic.

We found that there is a simple and efficient fine-grained control method, “Plug and Play Language Models (PPLM)”, introduced by Uber AI Research Institute. Easily allow large-scale language models to generate specific themed, stylized text, and have a wide range of applicability. In this paper, it took GPT-2 as an example to show how to control a big model. With Bayes’ rule, we have the equation:
$$p (x | a) \propto p (a | x) p (x)$$

\begin{figure}[ht]
\centering
\includegraphics[scale=0.3]{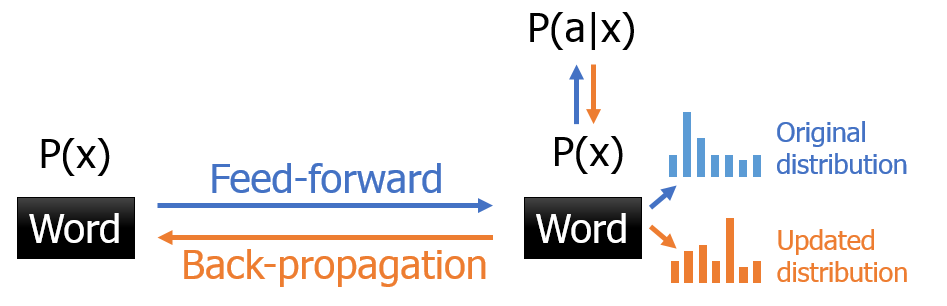}
\caption{the process of word generation towards desired attributes.}
\label{PPLM}
\end{figure}

Here, $p(x)$ is a probability distribution of language model over all the texts, regarding as the prior; $p (x | a)$ is the posterior, which is our goal that generating text with a certain attribute $a$; $p (a | x)$ is the bridge between what we have and what we want. According to the given sentences $x$, we can compute the probability of attribute $a$ in this $x$. It is small and easy to train. Through conducting feed-forward and back-propagation for the attribute model and the language model, we can get the gradient and use it to update parameters and sample words (Figure \ref{PPLM}).

Intuitively, as PPLM generates a word, its text representation gradually changes in the direction that more likely to meet the attribute requirements (large value of $p(a | x)$ ) and also maintaining the fluency of the original language model (large value of $p(x)$).

The particular direction comes from “Bag-of-words”, which is a txt file with several words. In other words, the more words appear in the generated sentences, the higher probabilities of the direction satisfaction.

By using the PPLM, we can, therefore, control the model to generate sentences in a particular direction. Here, our “Bag-of-words” can be three components: Topics, Storyline, and Nine relationships. For topics, we can select politics, computer, space, science, and so on. In terms of storyline, it is the keywords we extraction from specific scripts, and we will change based on the point we set before. In this way, we achieve our second highlight, which can push the plot development. Nine relationships are another critical part of our project, which is commonsense reasoning, and we will talk specifically.\\
\\
\textbf{Commonsense Reasoning.} People can easily infer the cause and effect of an event after observing a small segment of the event and even connect several events in causality. Due to the ability to reason about common sense, people can understand a long story like a person’s life or a country’s history or even the development of the universe through an only two-hour film. This ability is essential, which does not express well in most contemporary systems. As discussed above, the state-of-art model trained mainly based on a significant amount of data, which is depending on data but lack common sense.

ATOMIC, a commonsense knowledge graph distills inferential knowledge into nine different dimensions: person X’s intent, need, attribute, want, react, effect; others’ want, react, effect, respective corresponding to \textit{xIntent}, \textit{xNeed}, \textit{xAttr}, \textit{xWant}, \textit{xReact}, \textit{xEffect}, \textit{oWan}t, \textit{oReact}, \textit{oEffect}. Before this kind of form put forward, typical commonsense knowledge graphs are knowledge of ‘what’, such as OpenCyc, ConceptNet, EventNet and Formal Theory of Commonsense Psychology; while ATOMIC is the knowledge of ‘why’ and ‘how’, which is crucial because that’s how humans communicate and think.

COMET is a model that can automatically generate commonsense knowledge. In other words, given subject $s$ and relation $r$, it can generate object $o$. In the input part of the model, the researchers represented the triple $\{s, r, o\}$ as a sequence of words contained in each item of the triple, as shown below (X represents the input word):
$$\textbf{X} = \{X^S,X^T,X^O\}$$

Due to no need to consider the token order, location vector $p_t$ is put in each token. Moreover, COMET need to maximize the conditional likelihood of $X^O$:
$$\mathcal{L} = - \sum_{t=|s|+|r|}^{|s|+|r|+|o|} log P(x_t|x_{<t})$$

Training based on ATOMIC, it performs almost like a human. Hence, we use this model to generate Nine relationships and put them into “Bag-of-words” to improve the effect of text generation.\\
\\
So far, we have already built up the pipeline for playing our text game, and the next section will be personality classification. Before that, we would like to share some merits that we set to improve players’ experience.

\begin{itemize}
\item \textbf{Feel free to input text.} The player can text anything they want and even input nothing to show that they do not want to say something. Compared with other text games, which only give several choices and the player has to choose one, our game is just like the real chat with people in the story.
\item \textbf{Contextual coherence.} In our model, we always consider the latest talks like ten latest conversations so that the texts generated in a contextual coherence way.
\item \textbf{Mode optional.} In most of the text games, the character a player chosen only needs to react with several times though they appear many times. It is good for plot development and the same in our game. The player will be able to input texts after the character appears several times; we also can achieve another mode so-called ”immersive” which allows the player to control the character like immersing into the story.
\item \textbf{Universal.} We use the \textit{Friends} and \textit{Sherlock Series 1} as examples to show the process in our model, but actually, most of the scripts should be able to load in our model. The model just like a game console, and you can load different stories/games with different Cassettes.
\end{itemize}

\subsection{Personality classification}
\subsubsection{Dataset view}

To model the individual personality, we use MBTI (Myers-Briggs Type Indicator) by \textit{ C. G. Jung} as a measurement.  The identification of basic preferences of each of the four dichotomies specified or implicit in Jung's theory. Hence, we interact with four basic preferences to get 16 distinctive personality types. These four preferences are Favourite world (Outer/inner), Information (Sensing/Intuition), Decisions (Thinking/Feeling), and Structure (Judging/Perceiving).

Fortunately, we found a dataset with 8675 samples, where each sample contains 50 posts (separated by ‘$|||$') on SNS by a person and the resulting MBTI type. In this task, we mainly focus on training a model to classify the MBTI type.\\
\begin{figure}[H]
    \centering
    \includegraphics[scale=0.3]{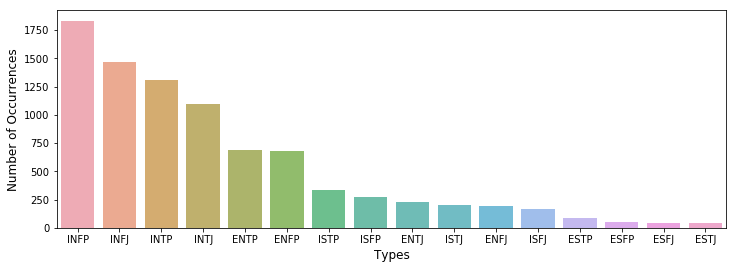}
    \caption{Frequency in each types}
\end{figure}

One thing is that this dataset is collected from a certain group of people. Therefore, people in this dataset tend to have a similar way of thinking, planning, and expressing. It suggests that training accuracy on those rare types might be lower and more varied.
\subsubsection{Data pre-processing}

As these posts are collected from SNS, there are two issues we have to solve before going into modeling. First is the gramma issue; most of these posts are not all formally wrote. Some abbreviations, such as np (no problem), tbh (to be honest), or btw (by the way), may not be embedded very well later on. Separately, there are also many URL links in these posts. Embedding on these links contributes very few to sentence representation.

Our pre-processing procedure first separation token ‘$|||$' by space. Next, we subtract all the URL links by ‘link’. The reason for this second step is that if we are deleting those links, the sentence will become incomplete. Then, we transfer words to lower case and remove unnecessary spaces. All the sentences are lemmatized by WordNetLemmatizer and removed their stopwords by NLTK package. At each step, we also transfer the string label of MBTI type into one-hot notation. 

\subsubsection{Classifiers}
We tried two simple approaches first, Multinomial Naive Bayes, and XGBoost. They are trained and compared under k-fold stratified validation.\\
\\
\textbf{Multinomial Naive Bayes.} This method comes from Bayesian' rule. The word 'Naive' indicates that this model is built on an ideal situation; features are assumed to be independent so that the relationship between features are not considered. This method was used on spam email detection. It conducts an association between spam and regular emails by choosing tokens (usually they are words in the text), and use Bayesian' rule to compute probabilities for classification.

In our task, we have 16 types of personalities. For personality type $j$, word $w$ at a word frequency of $f_w$:
$$Pr(j)\propto \pi _{j}\prod_{w=1}^{\left | V \right |}Pr(w|j)^{f_w}$$

In order to take stop words into account, we will add an Inverse Document Frequency (IDF) weight on each word:
$$t_w=(\frac{\sum_{n=1}^{N}doc_n}{doc_w})$$
$$Pr(j)\propto \pi _{j}\prod_{w=1}^{\left | V \right |}(t_wPr(w|j)^{f_w})$$

Even though the stop words have already been set to 0 for this specific use case, the IDF implementation is being added to generalize the function. \\
\\
\textbf{XGBoost.} Boosting produce a weak learner (classifier) in each step, weighted and added into the previous model. If the weak classifier in each step is generated according to the gradient of the loss function, it is Gradient Boosting. Suppose the optimal function is $F^{*}(x)$, i.e.:
$$F^{*}( \vec{x} )=\arg\min_{F} E_{(x,y)}[L(y, F( \vec{x}))]$$

Suppose $F(x)$ is the weighted sum of primary functions $f_i(x)$:
$$F(\vec{x})=\sum_{i=1}^{M}\gamma_i f_i(x)+ constant$$

We use the greedy algorithm in each step to choose the optimal solution.\\

\subsubsection{BERT: Pretrained-model }
Training in the BERT model has two steps, pre-training and fine-tuning. We use the pre-trained model by google and start from fine-tuning. The original BERT model for classification was fine-tuned for binary classification by adding a softmax function as the classifier. However, in our model, in order to perform a multilabel classification, we use the sigmoid function instead of softmax function. After training, we have the accuracy of each personality. 
As a result, we choose to use a classifier based on a pre-trained BERT model for MBTI classification. When generating process finishes, sentences will be processed by the same pre-processing method mentioned above. Then we choose the type with the largest probability to be the predicted MBTI type of the player and output the final summary for the user.

\section{Experiments, Results and Discussion}
\subsection{Plug and Play Language Model}
\begin{figure}[ht]
\centering
\includegraphics[scale=0.48]{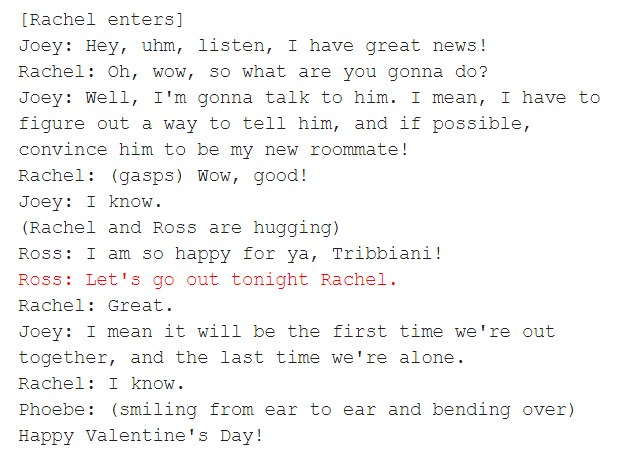}
\caption{partly result of text generation}
\label{eg1}
\end{figure}

When a player selects a story and a specific character, the game begins. Several texts will be given and the player can react by inputting texts. After several minutes (depend on the length of text generation), we get the response. The waiting time is a bit long since our game needs to considering several parts. Figure \ref{eg1} shows partly generation texts, and now the player is Ross and the sentence in color red is what he input. We can see that it seems quite reasonable roughly, those main characters talk one by one and there are also some comments or supplement sentences in brackets. However, when we look carefully, there are still some strange points. Furthermore, the generated texts cannot develop the plot well based on the storyline.
\

We have tried to figure out the better parameters for the model and also compared the improvement from the original GPT-2 to our new model. Though we take into account quite a lot of factors that may improve the results, it still cannot perform as we expect. We have three supposes and some of the points are trade-off, and we have to sacrifice accuracy for high speed:
\begin{itemize}
\item We only use the medium GPT-2 model.
\item The keywords we provide to the model is too general, and if we extract from every episode, it might be better.
\item We provide too much information in the “Bag-of-words”, and “Bag-of-words” might not be the best way to represent all these keywords.
\end{itemize}

\subsection{Personality Classification}
The final results give the accuracy of 0.5999 on Multinomial Naive Bayes and 0.6553 on XGB Classifier. The confusion matrix from  Multinomial Naive Bayes and XGB Classifier are shown in figure 5. The colour on the diagonal shows the degree of noise that the prediction has. We find that the personality type 'ESTJ' and 'ESFP' have more noise compared to other categories in both methods. We suppose the reasons for this are as follows.
\begin{itemize}
\item The dataset is imbalanced. The number of data in ‘ESFJ’ and ‘ESFP’ are both below 250.
\item The data in ‘ESFJ’ and ‘ESFP’ shows high variation such that the model can’t learn all the features in it.
\item The data in ‘ESFJ’ and ‘ESFP’ contains many link elements which is meaningless for classification.
\end{itemize}
.
\begin{figure}[h]
    \centering
    \includegraphics[scale=0.5]{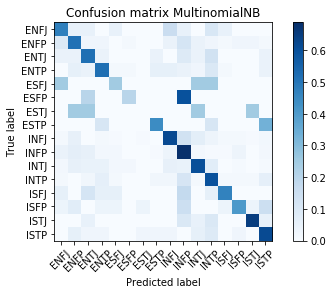}
    \includegraphics[scale=0.5]{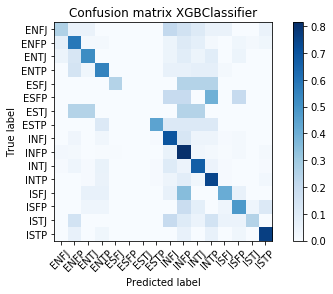}
    \caption{Confusion matrix}
    
\end{figure}\\

For the BERT model, we try different batch size such as 32, 64 and 128. As there is a trade off between the training speed and generalization of the model output, we choose a batch size of 32 to keep the generalization of the model and avoid the local minima. The loss of the model reached the global minima of 0.1786 at the step 487. The result given in table 1 is much better than the methods we used above. Most of the personality types have accuracy around 0.7. However, the same problem is shown here because of the imbalanced dataset, input variation and meaningless line elements. The accuracy of the type 'ENFJ' and 'ESFP' is much lower than the one of the others. 

\begin{table}[h]
\centering
\begin{tabular}{|l|l|}
\hline
     & Accuracy \\ \hline
INFJ & 0.6234   \\ \hline
ENTP & 0.7321   \\ \hline
INTP & 0.5931   \\ \hline
INTJ & 0.6762   \\ \hline
ENTJ & 0.6166   \\ \hline
ENFJ & 0.3074   \\ \hline
INFP & 0.7442   \\ \hline
ENFP & 0.6619   \\ \hline
ISFP & 0.6802   \\ \hline
ISTP & 0.7353   \\ \hline
ISFJ & 0.6986   \\ \hline
ISTJ & 0.7429   \\ \hline
ESTP & 0.6201   \\ \hline
ESFP & 0.5751   \\ \hline
ESTJ & 0.6421   \\ \hline
ESFJ & 0.7536   \\ \hline
\end{tabular}
\caption{Accuracy of the personality type}
\end{table}

Besides, there is also a mis-classification problem in BERT. Some kind of input such as ‘You are an idiot. I hate you stupid guy. Fine.’ might be classified into ‘ISTJ’, a serious, quiet and peace-living personality. This might be because we used a training sample with the length of 50 posts each time . Hence, as the new input is very short, there is a large probability that it will be mis-classified into ‘ISTJ’ compared to other categories. Even the largest probability of ‘ISTJ’ is just 0.2, which shows that it is closed to a random classification.

Lastly, other personalities, such as 'The Big Five' personality model can also be used for this task. The advantage in 'The Big Five' is it measures personality by a 5-dimension score vector. As a result, the numerical label vector might have better performance in predicting personalities.
\section{Conclusions}
In this paper, we present the Immersive Text game we built based on several methods related text generation, including PPLM, GPT-2 and COMET commonsense model.
The elaborate game mechanics and modes with the easily replaceable stories in the “cassette” form are designed in the game to enhance its entertainment and universality. 

The personality classification followed has good accuracy in most of the categories. However, the limitation in the dataset indicates the way of improving our model training in the future. We can choose a larger and more balanced dataset, sample from a more diverse range of people or make sentences more aligned with grammatical rules.

\section*{Acknowledgments}
We would like to offer our special thanks to Sebastian Riedel, Tim Rocktäschel and Pasquale Minervini, for their wonderful lectures on Natural Language Processing at UCL, leading us into this exciting area of research. We had a very enjoyable experience working together on the project, so thank you to all the team members for the interesting discussion and constant encouragement and support. In this very difficult period of time, we are very grateful to everyone for the continued efforts to keep up with the work and the good spirits, especially to Kexin and Jiaxuan who had to travel as advised for more than 20 hours.
Last but not least, thanks to our parents for their unfailing support and encouragement. Because of them, we can have a positive attitude towards difficulties and never give up, especially in this tough period.

\
\\

\bibliography{acl2019}
\bibliographystyle{acl_natbib}

\appendix


\section{Supplemental Material}
\subsection{Model comparison}
\begin{figure}[ht]
\centering
\includegraphics[scale=0.5]{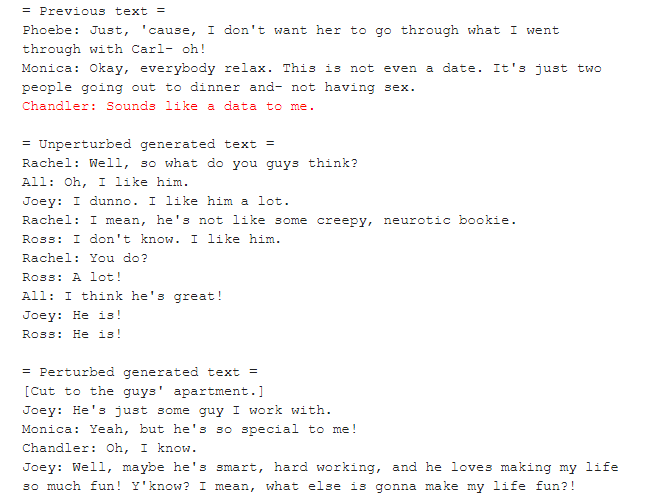}
\caption{improvement from GPT-2 to our new model.}
\label{PPLM_}
\end{figure}
Based on the previous text and the sentence we input (in color red), we can get the text generated from the GPT-2 and our model. We can see that through our model, the perturbed generated text is much better than the unperturbed one.

\subsection{How to play}
\begin{figure}[ht]
\centering
\includegraphics[scale=0.35]{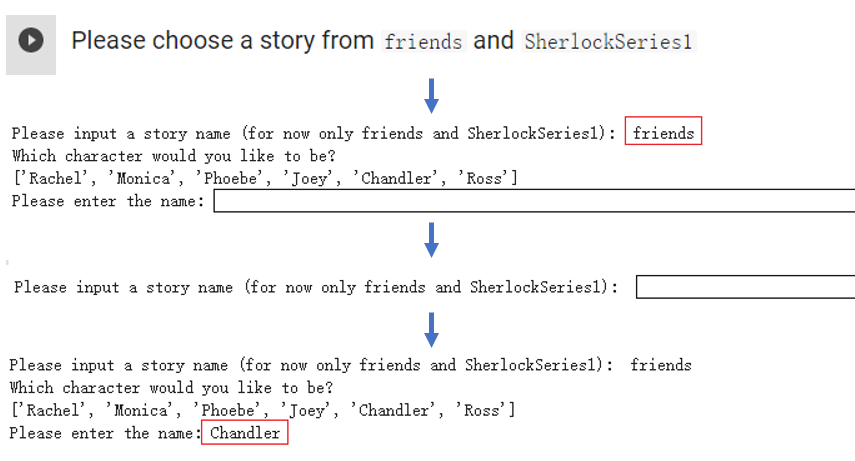}
\caption{choose a story and specify a character}
\label{e1}
\end{figure}

First of all, the player can choose a story and specify a character. Two available stories are available: friends and SherlockSeries1, and the option of corresponding characters in the stories will be displayed (Figure \ref{e1}).

\begin{figure}[ht]
\centering
\includegraphics[scale=0.33]{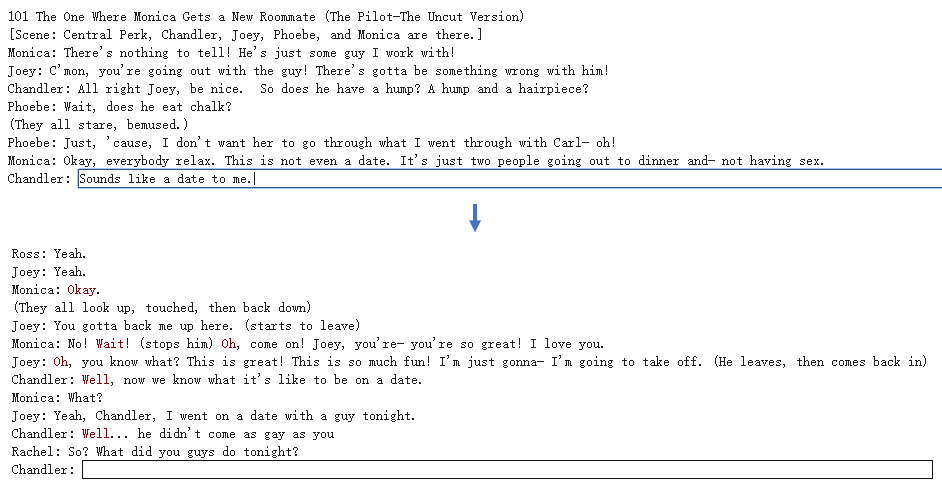}
\caption{interactive in the game}
\label{e2}
\end{figure}

According to the story the player chose, texts will display and the player can input what they want to say. After a while, new texts will be generated and the player can input again (\ref{e2}).

\newpage
\subsection{Description of personality types}
\
\begin{itemize}
\item \textbf{INFJ} The Protector: Quietly forceful, original, and sensitive. Tend to stick to things until they are done. Extremely intuitive about people, and concerned for their feelings. Well-developed value systems which they strictly adhere to. Well-respected for their perserverence in doing the right thing. Likely to be individualistic, rather than leading or following. 
\item \textbf{ENTP} The Visionary: Creative, resourceful, and intellectually quick. Good at a broad range of things. Enjoy debating issues, and may be into "one-up-manship" They get very excited about new ideas and projects, but may neglect the more routine aspects of life. Generally outspoken and assertive. They enjoy people and are stimulating company. Excellent ability to understand concepts and apply logic to find solutions.  
\item \textbf{INTP} The Thinker: Logical, original, creative thinkers. Can become very excited about theories and ideas. Exceptionally capable and driven to turn theories into clear understandings. Highly value knowledge, competence and logic. Quiet and reserved, hard to get to know well. Individualistic, having no interest in leading or following others.
\item \textbf{INTJ} The Scientist: Independent, original, analytical, and determined. Have an exceptional ability to turn theories into solid plans of action. Highly value knowledge, competence, and structure. Driven to derive meaning from their visions. Long-range thinkers. Have very high standards for their performance, and the performance of others. Natural leaders, but will follow if they trust existing leaders.
\item \textbf{ENTJ} The Executive: Assertive and outspoken - they are driven to lead. Excellent ability to understand difficult organizational problems and create solid solutions. Intelligent and well-informed, they usually excel at public speaking. They value knowledge and competence, and usually have little patience with inefficiency or disorganization.
\item \textbf{ENFJ} The Giver: Popular and sensitive, with outstanding people skills. Externally focused, with real concern for how others think and feel. Usually dislike being alone. They see everything from the human angle, and dislike impersonal analysis. Very effective at managing people issues, and leading group discussions. Interested in serving others, and probably place the needs of others over their own needs.
\item \textbf{INFP} The Idealist: Quiet, reflective, and idealistic. Interested in serving humanity. Well-developed value system, which they strive to live in accordance with. Extremely loyal. Adaptable and laid-back unless a strongly-held value is threatened. Usually talented writers. Mentally quick, and able to see possibilities. Interested in understanding and helping people.
\item \textbf{ENFP} The Inspirer: Enthusiastic, idealistic, and creative. Able to do almost anything that interests them. Great people skills. Need to live life in accordance with their inner values. Excited by new ideas, but bored with details. Open-minded and flexible, with a broad range of interests and abilities.
\item \textbf{ISFP} The Artist: Quiet, serious, sensitive and kind. Do not like conflict, and not likely to do things which may generate conflict. Loyal and faithful. Extremely well-developed senses, and aesthetic appreciation for beauty. Not interested in leading or controlling others. Flexible and open-minded. Likely to be original and creative. Enjoy the present moment. 
\item \textbf{ISTP} The Mechanic: Quiet and reserved, interested in how and why things work. Excellent skills with mechanical things. Risk-takers who they live for the moment. Usually interested in and talented at extreme sports. Uncomplicated in their desires. Loyal to their peers and to their internal value systems, but not overly concerned with respecting laws and rules if they get in the way of getting something done. Detached and analytical, they excel at finding solutions to practical problems.
\item \textbf{ISFJ} The Nuturer: Quiet, kind, and conscientious. Can be depended on to follow through. Usually puts the needs of others above their own needs. Stable and practical, they value security and traditions. Well-developed sense of space and function. Rich inner world of observations about people. Extremely perceptive of others feelings. Interested in serving others.
\item \textbf{ISTJ} The Duty Fulfiller: Serious and quiet, interested in security and peaceful living. Extremely thorough, responsible, and dependable. Well-developed powers of concentration. Usually interested in supporting and promoting traditions and establishments. Well-organized and hard working, they work steadily towards identified goals. They can usually accomplish any task once they have set their mind to it. 
\item \textbf{ESTP} The Doer: Friendly, adaptable, action-oriented. "Doers" who are focused on immediate results. Living in the here-and-now, theyre risk-takers who live fast-paced lifestyles. Impatient with long explanations. Extremely loyal to their peers, but not usually respectful of laws and rules if they get in the way of getting things done. Great people skills.
\item \textbf{ESFP}  The Performer: People-oriented and fun-loving, they make things more fun for others by their enjoyment. Living for the moment, they love new experiences. They dislike theory and impersonal analysis. Interested in serving others. Likely to be the center of attention in social situations. Well-developed common sense and practical ability.
\item \textbf{ESTJ} The Guardian: Practical, traditional, and organized. Likely to be athletic. Not interested in theory or abstraction unless they see the practical application. Have clear visions of the way things should be. Loyal and hard-working. Like to be in charge. Exceptionally capable in organizing and  running activities. Good citizens" who value security and peaceful living.
\item \textbf{ESFJ} The Caregiver: Warm-hearted, popular, and conscientious. Tend to put the needs of others over their own needs. Feel strong sense of responsibility and duty. Value traditions and security. Interested in serving others. Need positive reinforcement to feel good about themselves. Well-developed sense of space and function.
\end{itemize}

\end{document}